# Large Language Models in Medical Term Classification and Unexpected Misalignment Between Response and Reasoning

Xiaodan Zhang[1], Sandeep Vemulapalli[1,2], Nabasmita Talukdar[1], Sumyeong Ahn[1], Jiankun Wang[1], Han Meng[1], Sardar Mehtab Bin Murtaza[1], Aakash Ajay Dave[1], Dmitry Leshchiner[1], Dimitri F. Joseph[1], Martin Witteveen-Lane[2], Dave Chesla[1,2], Jiayu Zhou[1], and Bin Chen[1]

[1]Michigan State University, United States

[2]Corewell Health West Michigan, United States

*Keywords:* Mild Cognitive Impairment, Metastatic Cancer, Large Language Models, Model Inconsistency, Prompt Engineering

## ABSTRACT

*This study assesses the ability of state-of-the-art large language models (LLMs) including GPT-3.5, GPT-4, Falcon, and LLaMA 2 to identify patients with mild cognitive impairment (MCI) from discharge summaries and examines instances where the models' responses were misaligned with their reasoning. Utilizing the MIMIC-IV v2.2 database, we focused on a cohort aged 65 and older, verifying MCI diagnoses against ICD codes and expert evaluations. The data was partitioned into training, validation, and testing sets in a 7:2:1 ratio for model fine-tuning and evaluation, with an additional metastatic cancer dataset from MIMIC III used to further assess reasoning consistency. GPT-4 demonstrated superior interpretative capabilities, particularly in response to complex prompts, yet displayed notable response-reasoning inconsistencies. In contrast, open-source models like Falcon and LLaMA 2 achieved high accuracy but lacked explanatory reasoning, underscoring the necessity for further research to optimize both performance and interpretability.*

*The study emphasizes the significance of prompt engineering and the need for further exploration into the unexpected reasoning-response misalignment observed in GPT-4. The results underscore the promise of incorporating LLMs into healthcare diagnostics, contingent upon methodological advancements to ensure accuracy and clinical coherence of AI-generated outputs, thereby improving the trustworthiness of LLMs for medical decision-making.*

**BACKGROUND AND SIGNIFICANCE**

The advent of large language models (LLMs) such as OpenAI's GPT-3[1] and GPT-4[2] has revolutionized natural language processing (NLP), demonstrating human-like reasoning across diverse tasks. These models, trained on extensive and diverse datasets, are adept at uncovering intricate word relationships, leading to advanced capabilities in text generation and comprehension. A profound understanding of the connections between words can be applied to specialized fields like clinical informatics and applications that demand precision and confidentiality. For example, closed-source LLMs, particularly the GPT series developed by OpenAI, have demonstrated exceptional performance in various clinical tasks[3,4]. With their swift advancements, remarkably human-like performance, and powerful reasoning abilities, healthcare professionals are promoted to explore potential applications in real clinical settings. However, their closeness can create difficulties in effectively customizing them for specialized fields, and the characteristics that necessitate direct data input may also present privacy risks. In contrast, open-source models such as Meta's LLaMA[5,6] and Falcon[7,8] demonstrate immense promise for healthcare integration, circumventing data privacy issues and facilitating widespread adoption[9]. Whether the model is disclosed or not, there has been recent attention on studies involving prompt engineering to optimize input template formats. This is done to maximize the utilization of the reasoning capability of LLMs, especially in professional contexts. Strategic design of prompts can significantly enhance model output quality, especially in complex tasks demanding causal reasoning[10]. Recent studies demonstrate that utilizing structured prompts can elevate LLM performance in medical diagnostics, where precision and interpretability are essential[11,12,13].

Identifying patient comorbidities from unstructured clinical text is a routine challenge in clinical informatics. Extracted structural features from this text offer supplementary information to the recorded International Classification of Diseases (ICD) codes. One commonly misrepresented condition is Mild Cognitive Impairment (MCI), characterized by a noticeable decline in cognitive abilities, marking an intermediate stage between normal cognitive aging and more severe conditions like Alzheimer's disease (AD) and other types of dementia[14,15]. Early and accurate detection of MCI is essential for timely interventions. Applying LLMs to analyze clinical narratives, including patient discharge summaries, presents an innovative approach to MCI diagnosis, offering scalability, speed, and potentially greater accuracy than traditional methods like simple keyword matching. A variety of NLP tools have been explored to extract clinically relevant information from unstructured clinical notes in Electronic Health Records (EHRs). Various approaches involved the use of NLP tools to extract key concepts associated with cognitive function[16] and developing NLP-powered tools to streamline chart reviews for cognitive status determination[17]. Predictive models developed specifically for MCI identification, which operated in scenarios lacking screening scores, demonstrated moderate predictive accuracy[18]. Studies at Stanford Healthcare studies have developed models which successfully predicted MCI onset up to a year before formal diagnosis, underscoring the value of clinical factors derived from notes[19]. Additionally, research incorporating temporal characteristics of patient data into deep learning models showed enhanced predictive power for MCI[20]. A deep learning model based on keywords from clinical notes also demonstrated impressive performance in detecting MCI, although its use was restricted to a single dataset[21].

While these previous studies have demonstrated the potential of mining unstructured notes for MCI identification, our study takes a different approach. We aimed to use the identification of MCI patients from unstructured text as a case study, to specifically evaluate various prompting strategies and LLMs in terms of precision, interpretability, and privacy. To verify our findings, we further extend to metastatic cancer, another comorbidity not precisely presented by ICD codes[22]. We compare non-adapted ("zero-shot") and adapted ("fine-tuned") models. The non-adapted model is used directly for unfamiliar tasks/domains, while the adapted model is fine-tuned on training data to bridge distribution gaps. This highlight refining diagnostic methods and integrating LLMs into clinical workflows. Furthermore, we explore the critical issue of discrepancies between binary decisions made by LLMs and their accompanying rationale. Unlike the hallucination normally seen in GPT, these inconsistencies present unique challenges in clinical settings. Our thorough examination of GPT-4 in MCI and metastatic cancer identification emphasizes the need for models to produce outputs that are not only accurate but also logically coherent, a crucial aspect for clinical interpretability.

**MATERIALS AND METHODS**

**Dataset**

We conducted a retrospective cohort study using the MIMIC-IV v2.2 database[23], which houses de-identified health data from patients within critical care settings. This extensive dataset comprises records from 145,845 patients across 4,207,078 discharge summaries, offering a wealth of clinical narratives and diagnostic codes for analysis.

Inclusion criteria targeted individuals aged 65 and above, reflecting the common age range associated with the onset of MCI and dementia. Our initial MCI cohort was identified using ICD codes specific to MCI (ICD-9 code '331.83' and ICD-10 code 'G31.84'). To mitigate the potential confounding impact of multiple hospitalizations, we included only the first recorded discharge summaries for patients with recurrent admissions. Subsequently, we excluded any patients with ICD codes indicative of dementia or AD. A control group was constituted from the subset of patients 65 and above with the absence of ICD codes for MCI, dementia, or AD. From this pool, we randomly sampled 500 patients. A trio of medical experts reviewed the discharge summaries. Documentation that suggested MCI, either historical diagnosis or current assessment, was annotated for inclusion in the MCI group. Narratives documenting a transition from MCI to dementia were retained within the MCI group. Any instances of unrecognized MCI or dementia from the non-MCI group were excluded to avoid misclassification. The study flowchart is shown in **Figure 1.**

**Data Preprocessing**

Several steps were undertaken to clean and prepare discharge summaries for subsequent analysis. The content starting from the "*Chief Complaint*" in each summary was considered essential for mining, and therefore was extracted. This was followed by the removal of non-essential administrative text, placeholders, and discharge instructions to focus on critical relevant information. Medical abbreviations were expanded to full terms, and the text was standardized by converting to lowercase and removing punctuation.

For the effective training and evaluation of the models while also fine-tuning hyperparameters on the validation set, the dataset was partitioned into training, validation, and testing to ensure a thorough evaluation of the predictive models with a ratio of 7:2:1, respectively.

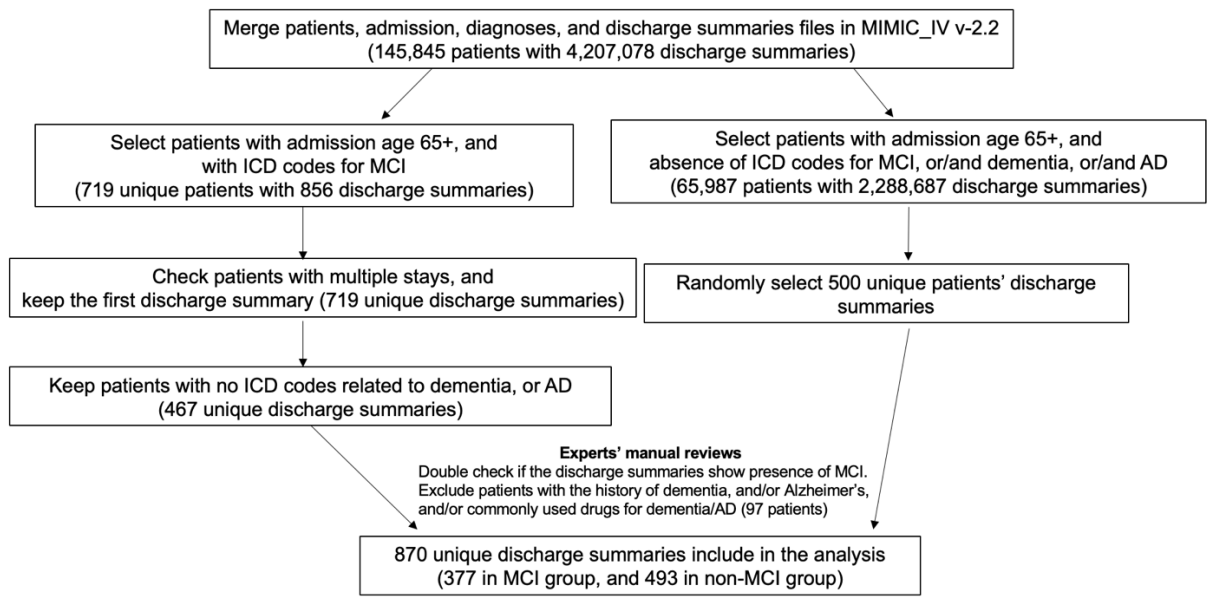

**Figure 1 Flowchart of the study.**

**LLMs**

A series of state-of-the-art LLMs, including GPT-3.5 turbo, GPT-4, Falcon, and LLaMA 2 were used in the zero-shot and/or fine-tuning approaches. The model was trained to recognize the mention of MCI in medical history and current conditions. We validated the model's performance against the expert-annotated dataset to assess its performance (i.e., F1 score, recall and precision). In our analysis, the "zero-shot" approach was exclusively applied to the testing set. The fine-tuning process encompassed the training, validation, and testing sets. To assess and compare models performance accurately, we focused on the outcomes on the testing set for all models and methods employed.

**GPT:** We used Microsoft Azure OpenAI service to access the GPT models to analyze clinical discharge summaries. We established a resource group to optimize resource management and operation within Azure's cloud environment. This resource group played a crucial role in systematically managing the computational assets required for the project. Following this, we developed a machine learning workspace, acting as a centralized hub for the construction and deployment of the GPT models. This setup enabled the efficient processing and interpretation of the discharge summaries. To address the rate limit issue encountered in GPT-4 due to high request frequency, we implemented intervals between consecutive requests and segmented the data into smaller batches.

**Meta LLaMA 2:** LLaMA 2 version 2 (V2) was released by Meta to the public in July 2023 with multiple variants of parameters (e.g., 7B, 13B, and 70B). For the zero-shot learning analysis, we utilized the 7B model accessed through the Hugging Face platform. The model was operated in a local environment, employing Langchain alongside other tools like scikit-learn and transformers for comprehensive processing. In the LLaMA 2-7B fine-tuning for the task, we implemented 4-bit floating-point quantization[24] and LoRA[25] for a more efficient fine-tuning process. These techniques allowed us to update only about 1.1% of the model's parameters (7 billion in total), with this percentage derived from the combination of the reduced parameter count due to LoRA's efficient adaptations and the additional compression from quantization. These methods collectively greatly enhanced efficiency, making the fine-tuning process more manageable and resource-efficient.

**Falcon**: We implemented the Falcon-7B model[26], a causal decoder-only model developed by the Technology Innovation Institute. Falcon-7B is accessible via the Hugging Face library and has undergone training on RefinedWeb, a meticulously curated dataset comprising over 1.5 trillion tokens derived from an expansive, deduplicated collection of web-based content. In our task, we first generated a classification model from the pre-trained Falcon model and then fine-tuned it using the training data. To evaluate the performance of the pre-trained model, we tested the generated classification model using the test data without fine-tuning. Most of the weights of the classification model are derived from the pre-trained model, with a small portion of the model, specifically the prediction layers, being randomly initialized. Similarly, we also applied 4-bit floating-point quantization and LoRA. Only about 0.9% of the parameters were updated during fine-tuning.

**Prompt Engineering**

In our investigation of LLMs for identifying MCI from discharge summaries, we utilized a series of prompts to explore various aspects of prompt engineering. Our exploratory process included a diverse range of prompts, from creative and complex structures to more direct and criteria-focused approaches. It became evident that while creative prompts have their appeal, their utility in medical diagnostic tasks is limited compared to approaches that provide clear instructions and adhere closely to diagnostic criteria. In addition, we explored some universal prompts such as "*Let's think*

*step by step*" and "*Take a deep breath*" and the results were quite comparable. However, our analysis revealed that prompts specifying a medical role, offering concise binary response instructions, and adhering to established MCI diagnostic criteria yielded the most significant results. We also noted that prompts encouraging a comprehensive review of the patient's medical history enhanced performance. This insight informed our prompt design, emphasizing the importance of thorough analysis in clinical diagnostics. Consequently, we refined our approach to include five prompts as detailed in **Supplementary Table 1**.

To refine our analysis, we introduced an additional instruction across all above prompts for the GPT-4 model: "*Please give concise reasonings. If the response does not logically follow from the reasoning, the response is flagged for review.*" This measure was implemented to ensure the clarity and consistency of the model's outputs. Discrepancies between the model's binary responses and its detailed reasoning were evaluated to assess the coherence and reliability of the model's reasoning abilities in the context of evidence-based clinical decision. We introduced "rationalization corrections" as a technique to align the binary responses ('Yes' or 'No') with the underlying reasoning. This process involves a critical review step where responses that exhibit a mismatch between the conclusion and the accompanying rationale are flagged. Subsequent corrections are applied to ensure that the final output is not only accurate but also logically consistent.

**Case Study: Metastatic Cancer Patient Classification Using Discharge Summaries**

To further examine the prevalence of inconsistencies observed in LLM outputs, our study expanded to a case study on the classification of metastatic cancer from the discharge summaries. We used the approach previously presented in[27] to prepare the metastatic cancer dataset in MIMIC-

III data[28]. A 7:2:1 ratio was implemented to split the data into training, validation, and testing sets. The same five prompts, originally crafted for MCI study, were adapted to the 188 testing set to discern the LLM's ability to detect and reason about the spread of cancer within discharge summaries (**Supplementary Table 2**). This extension served a dual purpose: to verify the model's diagnostic accuracy in a new medical domain and to scrutinize the consistency between the response and its reasoning. For example, Prompt 3, previously designed for a neurologist, was recontextualized for an oncologist's expertise in metastasis, while Prompt 5 was revised to outline metastatic cancer characteristics before soliciting the model's assessment. This case study examines if the challenge of achieving coherent and reliable outputs from LLMs are not exclusive to MCI but also manifest in other tasks.

**RESULTS**

From the MIMIC-IV v2.2 dataset, the selection process yielded 870 unique discharge summaries for the MCI dataset, with 377 summaries in the MCI group and 493 in the non-MCI group. In the metastatic cancer study, we used a MIMIC III dataset with 1,610 expert-annotated discharge summaries, of which 178 were identified as metastatic cancer cases.

**GPT 3.5 and GPT 4 Zero-shot**

Our comparative analysis of the zero-shot classification capabilities of GPT-3.5 Turbo and GPT-4 models revealed GPT-4's superior precision, especially notable in Prompt 4, which achieved a perfect recall and an F1 score of 0.964. GPT-4's score on Prompt 4 is substantially higher than

GPT-3.5 Turbo's 0.553 for the same prompt (**Table 1**). GPT-3.5 Turbo showed consistent performance across prompts, with a mean F1 of 0.564 (SD 0.056), while GPT-4's higher mean F1 of 0.857 (SD 0.112) indicates a more robust accuracy and a slightly greater variability in its classifications. This suggests that while GPT-4 has an increased capacity for understanding and responding to various prompts, the range of its performance is also broader, reflecting the model's flexible adaptation to diverse inputs.

**Table 1 Performance Metrics for GPT-3.5 turbo and GPT-4 Zero-shot Models**

| Models | Prompt | F1 | Recall | Precision |
|---|---|---|---|---|
| GPT-3.5 Turbo Zero-shot | Prompt 1 | 0.495 | 1.000 | 0.329 |
| | Prompt 2 | 0.593 | 1.000 | 0.422 |
| | Prompt 3 | **0.642** | 0.963 | **0.481** |
| | Prompt 4 | 0.553 | 0.963 | 0.388 |
| | Prompt 5 | 0.535 | 1.000 | 0.365 |
| | Mean (SD) | 0.564 (0.056) | **0.985 (0.020)** | 0.397 (0.058) |
| GPT-4 Zero-shot | Prompt 1 | 0.760 | 0.704 | 0.826 |
| | Prompt 2 | 0.714 | 0.556 | 1.000 |
| | Prompt 3 | **0.926** | 0.926 | 0.926 |
| | Prompt 4 | 0.964 | **1.000** | 0.931 |
| | Prompt 5 | 0.920 | 0.852 | 1.000 |
| | Mean (SD) | **0.857 (0.112)** | 0.808 (0.178) | **0.937 (0.071)** |

**LLaMA 2 Zero-shot & Fine-tuning**

In the LLaMA 2 zero-shot learning, interestingly, the highest F1 score was obtained in prompt 1 and there was a subsequent decrease in performance with each prompt, while the F1 score increased with subsequent prompts in the other models (**Table 2**). Overall, the highest F1 score was only 0.411, prompting us to investigate fine tuning methods for better results.

In LLaMA 2 fine-tuning, the number of training epochs is set to 10. To fully utilize the GPU memory, the training batch size for each GPU device is set to 2, with a gradient accumulation step of 4. The fine-tuned LLaMA 2 model showed superior performance over its zero-shot counterpart across all 5 prompts, emphasizing the significance of fine-tuning in identifying MCI. **Supplementary Figure 1** illustrates the performances under varying parameters, including the fine-tuning specifics for different numbers of epochs ranging from 1 to 10.

**Falcon Zero-shot & Fine-tuning**

In the Falcon zero-shot learning, the F1 scores of all prompts are lower than 0.5. Prompt 1, 2, 3 showed extremely high recall scores as most of the samples were predicted as positive, which makes it a clear sign of poor prediction capability (**Table 2**). Aligning with the observation in LLaMA 2, Falcon zero-shot learning appeared to be suboptimal, implying the limitation of in-context learning, and the necessity of fine-tuning for better performance.

Employing the same settings used in LLaMA 2 model, fine-tuned falcon models consistently performed better than zero-shot models. This reinforces the necessity and superiority of fine-tuning. Notably, more detailed instructions in the prompts led to better performance. For example, prompts 3 and 5, which offered detailed instructions, outperformed prompts 1 and 2. The best performance comes from prompt 3 with F1, recall, and precision scores of 0.971, 0.971, and 0.971. Although prompt 4 did not provide highly detailed instructions, it still achieved a high evaluation score. This may be attributed to its request for a '*comprehensive review*' of the notes, potentially prompting the model to engage in more thorough reasoning. More training details can be found in **Supplementary Figure 2**.

**Table 2 Performance Metrics for LLaMA 2 and Falcon zero-shot and fine-tuning models**

| Models | Prompt | Zero-Shot | | | Fine tuning | | |
|---|---|---|---|---|---|---|---|
| | | F1 | Recall | Precision | F1 | Recall | Precision |
| LLaMA 2 | Prompt 1 | **0.411** | **0.411** | **0.411** | 0.973 | 0.973 | 0.973 |
| | Prompt 2 | 0.383 | 0.370 | 0.383 | **1.000** | **1.000** | **1.000** |
| | Prompt 3 | 0.361 | 0.333 | 0.392 | 0.987 | 0.992 | 0.982 |
| | Prompt 4 | 0.331 | 0.304 | 0.363 | 0.986 | 0.981 | 0.992 |
| | Prompt 5 | 0.262 | 0.194 | 0.424 | 0.986 | 0.981 | 0.992 |
| | Mean | 0.350 | 0.322 | 0.395 | **0.986** | **0.985** | **0.988** |
| | (SD) | (0.057) | (0.082) | (0.024) | **(0.010)** | **(0.011)** | **(0.010)** |
| Falcon | Prompt 1 | 0.356 | 1.000 | 0.325 | 0.908 | 0.910 | 0.912 |
| | Prompt 2 | 0.345 | 1.000 | 0.321 | 0.780 | 0.780 | 0.785 |
| | Prompt 3 | 0.322 | 1.000 | 0.314 | **0.971** | **0.971** | **0.971** |
| | Prompt 4 | 0.448 | 0.407 | 0.256 | 0.960 | 0.960 | 0.960 |
| | Prompt 5 | **0.448** | **0.519** | **0.286** | 0.960 | 0.960 | 0.960 |
| | Mean | 0.384 | 0.785 | 0.300 | **0.916** | **0.916** | **0.918** |
| | (SD) | (0.060) | (0.297) | (0.029) | **(0.080)** | **(0.080)** | **(0.078)** |

Overall, fine-tuning demonstrates a clear advantage over zero-shot learning models. The LLaMa2 model with fine-tuning, achieved a robust mean F1 score of 0.986 and maintained a remarkably low standard deviation of 0.010, indicating both high performance and consistency across prompts. This contrast is even more pronounced when compared to its zero-shot learning performance, which had a significantly lower mean F1 score of 0.350 and a higher standard deviation of 0.057. Similarly, the Falcon model also showed improved results with fine-tuning, underlining the effectiveness of this approach for enhancing model precision in complex tasks.

**Inconsistencies in Logical Reasoning and Binary Predictions of GPT Models**

While we requested the GPT-4 model to provide binary responses along with reasoning, an unexpected pattern of significant inconsistency emerged. In other words, there was a discrepancy between the model's prediction and its verbal explanation. Predominantly, we observed a notable trend where the model's 'Yes' responses contradicted the reasoning that actually implied a 'No' prediction, while such a contradiction was less common in the model's 'No' responses. Some examples of these discrepancies are detailed in **Supplementary Table 3.**

Within the 87 testing set in the MCI dataset, we identified that prompt 1 and prompt 5 each displayed 10 (11.5%) instances where the 'Yes' predictions in the model contradicted with the reasonings that suggested a 'No' outcome, leading to inferior F1 performance (0.667 for prompt 1 and 0.727 for prompt 5) (**Table 3**). However, by implementing rationalization corrections to align the binary predictions with the associated reasoning, we achieved a significant improvement in F1 performance (0.788 for prompt 1 and 0.821 for prompt 5). This enhancement underscores the potential for increased accuracy when the model's output is in harmony with its reasoning process. Notably, prompt 4 showed remarkable consistency in its F1 score, maintaining high accuracy both before and after adjustments (0.852 to 0.912), highlighting the robustness of its predictive reliability.

We used Cohen's kappa κ to quantify the agreement between the model's predictions and the true labels. While the kappa values for actual predictions showed only a moderate agreement with the truth (0.439, 0.774, and 0.569 for prompts 1, 4, and 5, respectively), the reasoning-based predictions yielded higher kappa values, suggesting a stronger alignment with the true outcomes when the model's reasoning is correctly interpreted. This is most notably exemplified by prompt 4, which achieved a kappa value of 0.870, indicating a robust agreement.

**Table 3 Comparative Analysis of Response Accuracy for GPT-4 Models Before and After Rationalization Corrections in the MCI dataset**

| Prompts | n (%) | Actual Predictions | | | | Reasoning-Based Predictions | | | |
|---|---|---|---|---|---|---|---|---|---|
| | | κ | F1 | Recall | Precision | κ | F1 | Recall | Precision |
| Prompt 1 | 10 (11.5) | 0.439 | 0.667 | 0.963 | 0.510 | 0.665 | 0.788 | 0.963 | 0.667 |
| Prompt 2 | 2 (2.3) | 0.726 | 0.808 | 0.778 | 0.840 | 0.776 | 0.840 | 0.778 | 0.913 |
| Prompt 3 | 0 (0) | 0.742 | 0.828 | 0.889 | 0.774 | 0.742 | 0.828 | 0.889 | 0.774 |
| Prompt 4 | 4 (4.6) | 0.774 | **0.852** | 0.963 | 0.765 | **0.870** | **0.912** | 0.963 | 0.867 |
| Prompt 5 | 10 (11.5) | 0.569 | 0.727 | 0.889 | 0.615 | 0.737 | 0.821 | 0.852 | 0.793 |

*n* (%): Inconsistencies counts and percentage; κ: Cohen's kappa with the true labels

In the subsequent case study using the metastatic cancer dataset, similar patterns of inconsistency were observed among the 188 testing set. Applying the rationalization corrections again resulted in a significant improvement in performance. For example, prompt 5 had a significant enhancement from 0.552 to 0.800 after rationalization (**Table 4**). These enhancements confirm the potential of rationalization in refining the predictive accuracy of LLMs. Cohen's kappa values also revealed a notable improvement post-rationalization for prompt 4, indicating a more robust agreement with the true labels. These findings mirror the patterns observed in the MCI data, further emphasizing the importance of careful examination and adjustment of LLM outputs in medical settings.

**Table 4 Comparative Analysis of Response Accuracy for GPT-4 Models Before and After Rationalization Corrections in the metastatic cancer dataset**

| Prompts | n (%) | Actual Predictions | | | | Reasoning-Based Predictions | | | |
|---|---|---|---|---|---|---|---|---|---|
| | | κ | F1 | Recall | Precision | κ | F1 | Recall | Precision |
| Prompt 1 | 7 (3.7) | 0.384 | 0.467 | 0.778 | 0.333 | 0.453 | 0.520 | 0.722 | 0.406 |
| Prompt 2 | 3 (1.6) | **0.665** | 0.700 | 0.778 | 0.636 | 0.716 | 0.743 | 0.722 | 0.765 |
| Prompt 3 | 3 (1.6) | 0.628 | 0.667 | 0.722 | 0.619 | 0.716 | 0.743 | 0.722 | 0.765 |
| Prompt 4 | 15 (8) | 0.516 | 0.577 | 0.833 | 0.441 | **0.782** | 0.649 | 0.667 | 0.632 |
| Prompt 5 | 18 (9.6) | 0.484 | 0.552 | 0.889 | 0.400 | 0.776 | 0.800 | 0.889 | 0.727 |

*n* (%): Inconsistencies counts and percentage; κ: Cohen's kappa with the true labels

**GPT-4's Advanced Reasoning Capabilities**

The deployment of GPT-4 for MCI identification emphasized the model's advanced reasoning abilities, which surpass simple keyword matching. GPT-4 conducts a detailed examination of clinical narratives, pinpointing and interpreting relevant data points such as memory difficulties, orientation issues, and daily functioning impediments. It adeptly identifies patterns and makes connections that suggest cognitive impairment, considering the clinical context of MCI-related terms within a patient's history. The model's evaluation process is characterized by an intricate interpretation of symptoms and their progression, alongside cognitive performance markers, mirroring the discernment typical of human clinical reasoning. For instance, GPT-4 has been adept at detecting signs of MCI such as increased memory challenges, fluctuating cognitive function, and dependence on cues for recall, despite some retention of memory (**Supplementary Figure 3**). This thorough analysis, considering the dynamic nature of symptoms and their influence on daily life, showcases GPT-4's profound reasoning capability. It showed that our methodological approach not only identifies occurrences of MCI but also covers a wide spectrum of cognitive health conditions, providing a powerful tool for in-depth exploration of MCI's manifestation and its detailed documentation within medical records.

**DISCUSSION**

Our investigation into the application of LLMs for clinical term classification has unearthed significant insights into the art of prompt engineering. It became evident that prompts that outlined specific roles and provided methodical instructions yielded consistently higher results across GPT models, a finding that aligns with and is supported by existing literature on the significance of role-focused prompt design[29]. The phraseology within prompts is consequential; for example, "*your duty is to assess*," "*determine*" and "*conduct a comprehensive review*" has emerged as pivotal, suggesting the impact of precise language in soliciting accurate and clinically useful responses from the models.

The high F1 scores in Falcon and LLaMA 2 fine-tunings demonstrate their efficiency in binary classification, particularly when training and test data are well-matched. However, their use in clinical environments, where understanding the reasoning behind predictions is crucial, might be constrained. Although these models excel in delivering remarkable outcomes, GPT models stand out by also providing explanations, which are vital for clinical validation. Nevertheless, because of the concern about using OpenAI's API for processing clinical notes, open-source models remain promising in future clinical applications.

This discrepancy reveals the need for a more rigorous review of LLM outputs, particularly in nuanced clinical decision-making domains. Although LLMs can efficiently generate vast amounts of textual data for any inquiry, medical professionals are critical to ensuring clinical relevance and accuracy of the outputs. In most instances, the model tended to predict a positive response even when the actual response was negative. This discrepancy became evident when probing the model for reasoning, as it often leaned towards a negative prediction, revealing a notable inconsistency between the model's initial positive response and its subsequent reasoning. There are several factors that could contribute to the high false positive rate. For instance, some mismatches could arise from simple errors in the model's interpretation or processing of the data. In other cases, the reasoning suggests a potential or future risk for a condition like MCI, rather than a current diagnosis, leading to a 'Yes' prediction despite insufficient evidence. While in some cases, the task is so simple that a professional would never make a mistake, LLM unexpectedly rendered unreasonable responses. However, the reasoning normally makes sense. Given the prevalence of such inconsistencies between reasoning and response, it is recommended to conduct a thorough evaluation of the reasoning process rather than simply adopting the responses.

The study also emphasizes the need for further research and refinement in the application of LLMs in healthcare, particularly in improving the alignment between response and reasoning. The model may be improved by further training or fine-tuning using datasets that pair binary decisions with their justifications. The reliability of conclusions could also be improved by implementing hybrid systems that combine LLMs with rule-based logic. In addition, we propose developing human-LLM collaborative frameworks where medical experts review and corroborate LLM's initial screening. This can not only enhance decision-making accuracy but also maintain the transparency and interpretability essential for clinical applications. The adoption of ensemble methods, where multiple models are used to cross-verify results, could mitigate the inconsistency issue.

Several limitations should be considered when interpreting our results. First, this study was restricted to retrospective analysis and did not include prospective clinical validation. Although our findings are evident, the focus on MCI and metastatic cancer may limit the generalizability of findings across other medical conditions. Future research must address these limitations by exploring broader datasets and conducting more prospective analysis to validate the efficacy of LLMs in practice and explore the development of LLMs that integrate robust rationalization processes to ensure their outputs are clinically coherent.

**CONCLUSION**

In conclusion, our research underlines the effectiveness of fine-tuning over zero-shot learning when employing LLMs like LLaMA 2 and Falcon for clinical narrative analysis. Despite achieving high precision, those models lack effective explanatory reasoning, pointing to an urgent need for research focused on achieving both high performance and interpretability. The study also highlights the crucial need for rationalization corrections to address the misalignment between binary predictions and their underlying reasoning, a challenge exemplified by the advanced yet inconsistent reasoning capabilities in GPT-4. Further investigations into prompt engineering and the development of hybrid human-LLM collaborative frameworks are essential to enhance the interpretability and reliability of LLM outputs in healthcare applications, ensuring accuracy and logical consistency in model predictions.

**FUNDING**

The research is supported by the MSU-Corewell Health Alliance Funds and the NIH R01GM145700, R01GM134307, and 1RF1AG072449. The content is solely the responsibility of the authors and does not necessarily represent the official views of the funders.

**COMPETING INTERESTS**

The authors declare no competing interests.

**AUTHOR CONTRIBUTIONS**

X.Z. and S.V. were responsible for the design of the study and the preparation of datasets. M.W. assisted with dataset preparation. X.Z. took the lead in coordinating the research activities and implementing the GPT models. The development of prompts and exploration of prompt engineering were jointly conducted by N.T. and X.Z. S.V. was responsible for the exploration of LLaMA 2 in a zero-shot context. The LLaMA 2 fine-tuning was handled by S.A. and J.W.. H.M. diligently conducted experiments focusing on Falcon zero-shot and fine-tuning. S.M. participated in exploring the Falcon model. A.D., D.L., and D.J. participated in the manual annotation process. B.C., J.Z. and D.C. supervised the study and provided the resources. All authors actively participated in the preparation of the manuscript.

# Supplementary Files

**Table 1. Prompt Used in the Models Comparison for MCI**

**Table 2 Prompt Used in the Metastatic Cancer Dataset**

**Table 3 Illustrative Cases of Discrepancies Between the Binary Diagnoses and Underlying Reasoning in GPT-4**

**Figure 1 Training Loss, Validation Loss, and Test Accuracy Across Epochs During Fine-Tuning of LLaMA 2 Models**

**Figure 2 Training Loss, Validation Loss, and Test Accuracy Across Epochs During Fine-Tuning of Falcon Models**

**Figure 3 Example of Discharge Summaries for a Patient with MCI (Key Factors Leading To Correct Identification By GPT-4 Model)**

**Table 1 Prompt Used in the Models Comparison for MCI**

| No. | Details |
|---|---|
| **Prompt 1** | Review this patient's discharge summary and consider the possibility of mild cognitive impairment. Give a 'Yes' or 'No' response. |
| **Prompt 2** | As a healthcare professional, please analyze the following discharge summary and determine if the patient has mild cognitive impairment. Please provide a concise 'Yes' or 'No' response. |
| **Prompt 3** | As a neurologist, your duty is to analyze each discharge summary and determine whether a patient has mild cognitive impairment. Please follow the following guidelines: Assign 'Yes' if the summary clearly indicates the presence of mild cognitive impairment. Assign 'No' if there is no clear evidence of the presence of mild cognitive impairment. |
| **Prompt 4** | Conduct a comprehensive review of the patient's documented medical journey in the discharge summary. Can you confirm the presence of mild cognitive impairment at any stage? Provide a 'Yes' or 'No' response. |
| **Prompt 5** | Review the patient's discharge summary. Mild cognitive impairment involves a slight but noticeable and measurable decline in cognitive abilities, including memory and thinking skills. Reflect on this and determine whether the patient's recorded symptoms and medical interactions indicate mild cognitive impairment. Respond with 'Yes' if the criteria are met, or 'No' if they are not. |

**Prompt 1:** Serves as a basic evaluation prompt, acting as a baseline for comparison.

**Prompt 2:** Introduces a medical role, resulting in the development of a healthcare professional evaluation prompt.

**Prompt 3:** Tailors the prompt for neurologists, with clear instructions for a 'Yes' or 'No' response based on MCI presence or absence were incorporated.

**Prompt 4:** Encourages a comprehensive review of the patient's medical journey, aiming to enhance interpretive depth. This is exemplified in the comprehensive review prompt.

**Prompt 5:** Adheres to diagnostic criteria in the diagnostic criteria-based prompt, including the explicit definition of MCI.

**Table 2 Prompt Used in the Metastatic Cancer Dataset**

| No. | Details |
|---|---|
| **Prompt 1** | Review this patient's discharge summary and consider the possibility of metastatic cancer. Give a 'Yes' or 'No' response. Please give concise reasonings. If the response does not logically follow from the reasoning, the response is flagged for review. |
| **Prompt 2** | As a healthcare professional, please analyze the following discharge summary and determine if the patient has metastatic cancer. Please provide a concise 'Yes' or 'No' response. Please give concise reasonings. If the response does not logically follow from the reasoning, the response is flagged for review. |
| **Prompt 3** | As an oncologist, your duty is to analyze each discharge summary and determine whether a patient has metastatic cancer. Please follow the following guidelines: Assign 'Yes' if the summary clearly indicates the presence of metastatic cancer. Assign 'No' if there is no clear evidence of the presence of metastatic cancer. Please give concise reasonings. If the response does not logically follow from the reasoning, the response is flagged for review. |
| **Prompt 4** | Conduct a comprehensive review of the patient's documented medical journey in the discharge summary. Can you confirm the presence of metastatic cancer at any stage? Provide a 'Yes' or 'No' response. Please give concise reasonings. If the response does not logically follow from the reasoning, the response is flagged for review. |
| **Prompt 5** | Review the patient's discharge summary. Metastatic cancer refers to cancers with very high mortality rate or imminent mortality (pancreas, esophagus, stomach, cholangiocarcinoma, and brain). This also covers situations where there is mention of distant or multi-organ metastasis, where palliative care would be considered (prognosis < 6 months). Reflect on this and determine whether the patient's recorded symptoms and medical interactions indicate metastatic cancer. Respond with 'Yes' if the criteria are met, or 'No' if they are not. Please give concise reasonings. If the response does not logically follow from the reasoning, the response is flagged for review. |

**Table 3 Illustrative Cases of Discrepancies Between the Binary Diagnoses and Underlying Reasoning in GPT-4**

| Response Outputs from GPT-4 Model |
|---|
| **Discharge Summary Case 1**<br><br>**Response: Yes**<br><br>**Reasoning:**<br>The patient's discharge summary **does not** indicate symptoms or signs of mild cognitive impairment. The neurologic assessment mentions that the patient is "awake, alert, and oriented times three"," with cognitive functions (CNS II-XII) " and "grossly intact." These findings suggest no current evidence of cognitive deficits, which would be typical in mild cognitive impairment. |
| **Discharge Summary Case 2**<br><br>**Response: Yes**<br><br>**Reasoning:**<br>The discharge summary **does not** indicate any specific cognitive issues, memory problems, disorientation, or difficulty with communication that are typical symptoms of mild cognitive impairment (MCI). The patient's mental status is described as "clear and coherent " and his level of consciousness as "alert and interactive" which suggests that there are no overt signs of cognitive impairment based on this document. |
| **Discharge Summary Case 3**<br><br>**Response: Yes**<br><br>**Reasoning:**<br>The patient's recorded symptoms and medical interactions indicate a neurological event, more precisely a subacute ischemic stroke of the right thalamus, rather than being indicative of mild cognitive impairment as the underlying condition. Although there are aspects of the neurological examination that assess cognitive function (such as the patient being alert, oriented, having fluent language with intact repetition, comprehension, and no evidence of apraxia or neglect), these findings were more focused on evaluating the effects of the recent stroke and ensuring there was no significant cognitive decline associated with it. The documentation states the patient was able to relate history without difficulty, following commands, and exhibiting no signs of language deficits or cortical sensory loss. This indicates that her cognitive faculties were intact post-stroke. |

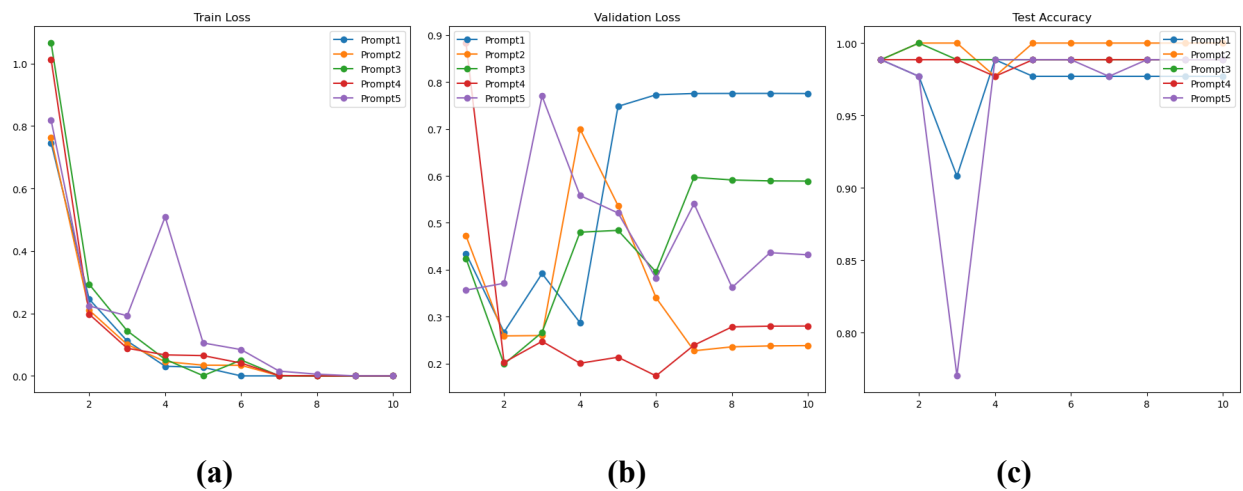

(a)                          (b)                          (c)

**Figure 1 Training Loss, Validation Loss, and Test Accuracy Across Epochs During Fine-Tuning of LLaMA 2 Models**

**Train Loss (a)**: All models start with high loss values. However, they quickly improve, with the loss dropping significantly by the second epoch. By the end of the 10 epochs, most models converge to a loss value near 0, except for Prompt5 showing a spike at epoch 7, which indicates a potential issue at that epoch, such as a bad batch of data or an anomaly in training.

**Validation Loss (b)**: The validation loss trends are more erratic compared to the training loss, with several peaks and valleys. This could be a sign of model overfitting to the training data, instability in the model.

**Test Accuracy (c)**: The test accuracy is relatively high for all models, mostly above 90%. There is, however, a noticeable dip for Prompt5 around the 3rd epoch, which could suggest a momentary degradation in its ability to generalize. By the 10th epoch, the accuracy of the models stabilizes, indicating that they have learned from the fine-tuning process.

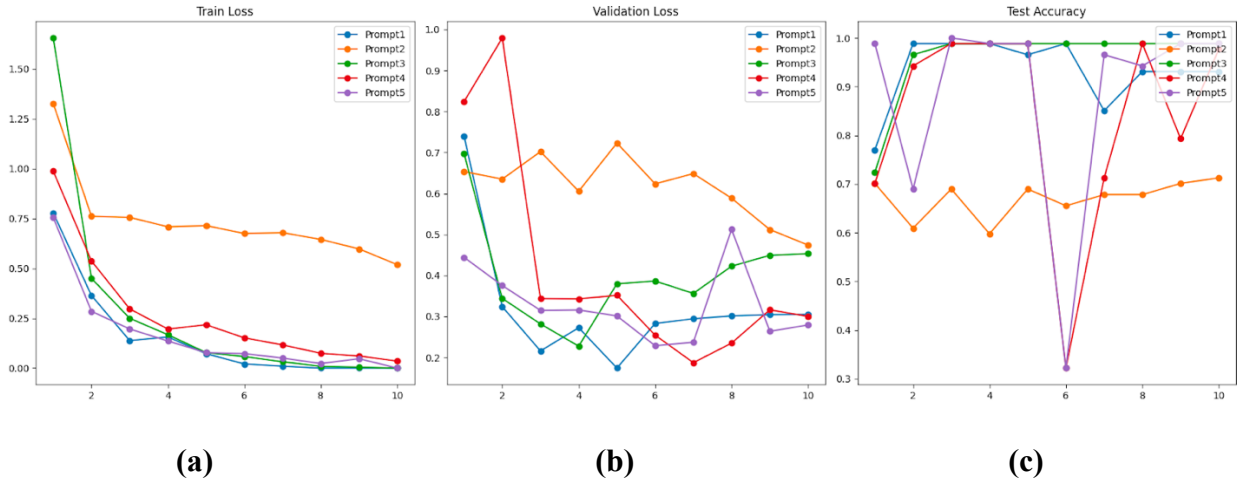

|  (a)  |  (b)  |  (c)  |

**Figure 2 Training Loss, Validation Loss, and Test Accuracy Across Epochs During Fine-Tuning of Falcon Models**

**Train Loss (a):** All models can quickly fit the training dataset and quickly converge. For most of the prompts, training losses are decreasing to nearly zero. For prompt 2, the loss only falls to about 0.5.

**Validation Loss (b):** Compared with training loss, there is more zig-zag in the validation loss curve. Only prompt 3 shows a clear overfitting sign after 4 epochs.

**Test Accuracy (c):** For prompts 4 and 5, there appears a deep valley in the curve. And for prompt 2, the test accuracy did not improve a lot during the finetuning.

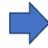

**Figure 3 Example of Discharge Summaries for a Patient with MCI (Key Factors Leading to Correct Identification by GPT-4 Model)**